\documentclass[letterpaper]{article} 
\usepackage{aaai24}  
\usepackage{times}  
\usepackage{helvet}  
\usepackage{courier}  
\usepackage[hyphens]{url}  
\usepackage{graphicx} 
\usepackage{natbib}  
\usepackage{caption} 
\usepackage{algorithm}
\usepackage{amsmath}
\usepackage{amssymb}
\usepackage{algpseudocode}
\usepackage{newfloat}
\usepackage{listings}
\usepackage{bm}
\usepackage{booktabs}

\usepackage{newfloat}
\usepackage{listings}
\DeclareCaptionStyle{ruled}{labelfont=normalfont,labelsep=colon,strut=off} 
\floatstyle{ruled}
\newfloat{listing}{tb}{lst}{}
\floatname{listing}{Listing}
\title{Scalable Motion Style Transfer with Constrained Diffusion Generation}
\author{
    Wenjie Yin\textsuperscript{\rm 1}, 
    Yi Yu\textsuperscript{\rm 2\thanks{Corresponding authors}}, 
    Hang Yin\textsuperscript{\rm 3}, 
    Danica Kragic\textsuperscript{\rm 1}, 
    Mårten Björkman\textsuperscript{\rm 1\footnotemark[1]}
}
\affiliations{
    \textsuperscript{\rm 1}KTH Royal Institute of Technology, Sweden\\
    \textsuperscript{\rm 2}National Institute of Informatics, Japan\\
    \textsuperscript{\rm 3}University of Copenhagen, Denmark\\
    \{yinw, dani, celle\}@kth.se, yiyu@nii.ac.jp, hayi@di.ku.dk
}

\begin{document}

\maketitle

\begin{abstract}
Current training of motion style transfer systems relies on consistency losses across style domains to preserve contents, hindering its scalable application to a large number of domains and private data. Recent image transfer works show the potential of independent training on each domain by leveraging implicit bridging between diffusion models, with the content preservation, however, limited to simple data patterns. We address this by imposing biased sampling in backward diffusion while maintaining the domain independence in the training stage. We construct the bias from the source domain keyframes and apply them as the gradient of content constraints, yielding a framework with keyframe manifold constraint gradients (KMCGs). Our validation demonstrates the success of training separate models to transfer between as many as ten dance motion styles. Comprehensive experiments find a significant improvement in preserving motion contents in comparison to baseline and ablative diffusion-based style transfer models. In addition, we perform a human study for a subjective assessment of the quality of generated dance motions. The results validate the competitiveness of KMCGs. 
\end{abstract}

\section{Introduction}
Human motion style is a complex and important aspect of human behavior, which can be perceived as motion characteristics that convey personality and temper or articulate socio-cultural factors. The ability to accurately capture and replicate motion style is crucial in numerous applications, such as video games, and choreography. 
Style transfer systems can streamline creation of various media, including images~\cite{gatys2016image}, music~\cite{brunner2018symbolic}, and indeed human movements~\cite{dong2020adult2child}. For instance, in the realm of video games, unique modes of action correspond to various player operations and character states. Similarly, in the field of dance choreography, each dance genre has its own movement patterns, and the application of style transfer can assist choreographers in creating variations of particular movements.

\begin{figure}[ht]
  \includegraphics[width=\linewidth]{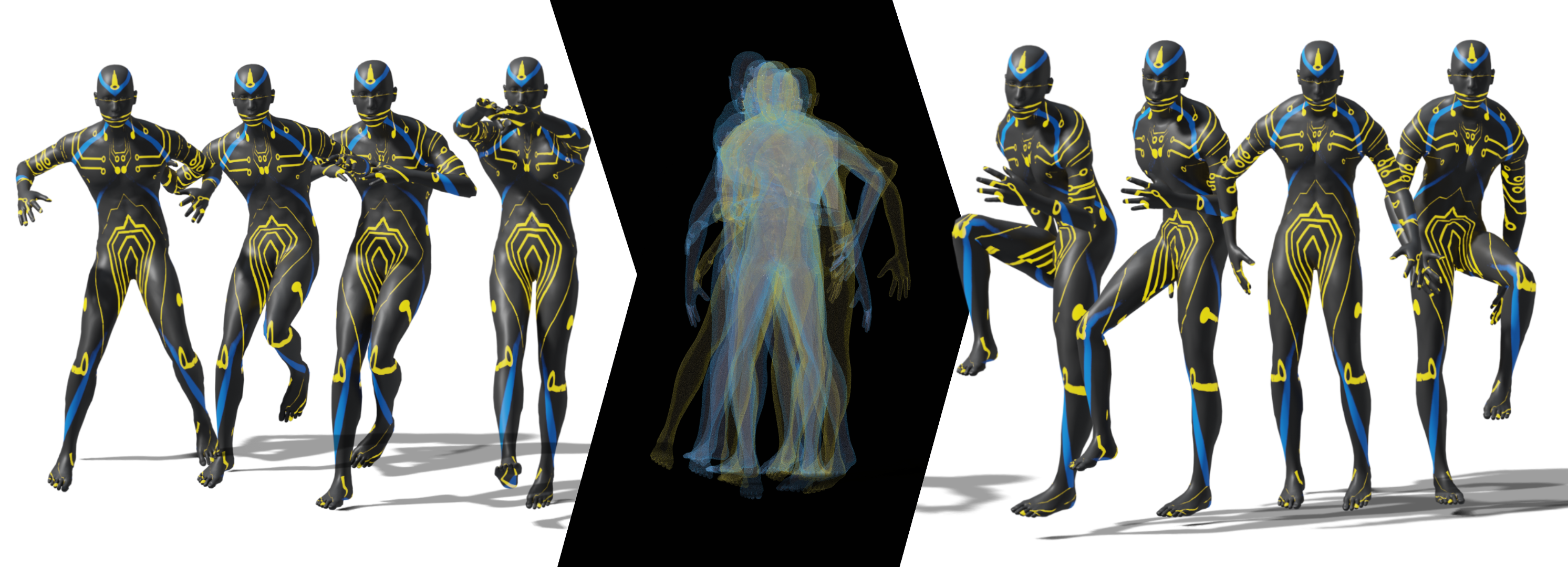}
  \caption{The proposed scalable motion style transfer system, with independently trained diffusion models. Given a source motion sequence, our system converts it to latent encoding, then constructs motion in the target style. }
  \label{fig:ddib_motion}
\end{figure}

Early methods for motion style transfer~\cite{hsu2005style, taylor2009factored, smith2019efficient} primarily relied on supervised learning, requiring paired and annotated data. However, obtaining matched human motion sequences is challenging and often involves a laborious preprocessing phase. Consequently, the predominant style transfer methods today are unpaired translations that do not demand direct mappings between motion samples. Common methods among these are the motion transfer systems based on CycleGAN~\cite{dong2020adult2child, yin2023dance} for paired style domains and StarGAN~\cite{chan2020emotion, yin2023multimodal} for a set of domains. 

While these systems are capable of generating high-quality human motion, they exhibit considerable limitations when it comes to scalability towards new domains. Specifically, CycleGAN-based methods are trained on distinct pairs of motion styles, where each model is tailored for transfer between a specific pair of styles~\cite{dong2020adult2child, yin2023dance}, or establish a shared domain bridging multiple styles via StarGAN-based architecture~\cite{chan2020emotion, yin2023multimodal}. 

The former implies training a quadratically increasing number of models, making it an impractical solution when transfer between a large number of style domains is considered. The latter faces finding such a shared domain, which is an enormously challenging task. Both methods also show limited incremental scalability and adaptability; a new style requires retraining of the models on the entire dataset. The requirement of simultanous access to the entire dataset highlights an additional drawback of these methods when parts of the data are sensitive due to privacy concerns, such as in rehabilitation therapy. Developing methods that address these prominent challenges is thus critical for building practical motion style transfer systems.

In this paper, we propose a method based on Dual Diffusion Implicit Bridges (DDIBs)~\cite{su2022dual} to mitigate both issues of system scalability and data privacy. DDIBs adopt the Schrödinger bridge perspective of diffusion models, showing certain information can be passed between diffusion latent spaces and used for image translation. Our system leverages this for a motion style transfer system with a two-step process. Given a source and a target model, the system initially utilizes the source model to obtain a latent encoding of motion at the final diffusion time step. This latent encoding is subsequently fed as the starting condition to the target model to generate the target motion (See Figure \ref{fig:ddib_motion}). The source and target domain models are completely decoupled, which allows for training the models separately. As a result, our system alleviates the need for a paired dataset that is typical for CycleGAN- and StarGAN-based methods, while upholding data privacy.

Our core technical contribution is improving content preservation of DDIBs in the motion style transfer domain. We find DDIBs struggle to retain the motion content faithfully when the analogy between source and target domains is low, consistent with the observations in image translation~\cite{su2022dual}. This is probably due to uncontrolled information encoding and decoding in independent diffusion processes. To this end, we propose a Keyframe Manifold Constraint Gradients (KMCGs) framework to improve content coherence of the target domain during inference. 
KMCGs uses keyframes from the source domain as context constraints and employ Manifold Constrained Gradient~\cite{chung2022improving} to enforce these constraints during the second phase of DDIBs. Our experiments and quantative analysis on the 100STYLE~\cite{mason2022real} locomotion database, and the AIST++~\cite{tsuchida2019aist} dance database find KMCGs achieves successful style transfer and better content preservation, reflected on the cycle consistency property on these two motion datasets and two probabilistic divergence-based metrics. Additionally, our human study as a subjective evaluation find samples generated from KMCGs are preferred in general. Overall, these evaluations demonstrate that our system significantly outperforms baseline methods. 
A video of the summary and examples can be accessed via \url{https://youtu.be/98wEWavjnxI}.

In summary, our contributions are:
\begin{itemize}
\item A motion style transfer system that generates motions ranging from fundamental human locomotion to sophisticated dance movements. Our system pioneers in terms of system scalability and data privacy, demonstrating efficient and independent training over ten styles. 
\item A technical method KMCGs that mitigates the content coherence issue of dual diffusion implicit bridges, which is found prominent in transferring complex motions. 
\item 
A comprehensive evaluation of the proposed style transfer system including both objective metrics and subjective human study, showing significant performances boost compared to baseline and ablative models.
\end{itemize}

\section{Related Work}
\label{sec: related work}
We now review relevant prior works, including diffusion-based motion synthesis, as well as motion style transfer. 

\subsection{Diffusion-based Motion Synthesis}
\label{subsec: diffusion-based motion synthesis}

In light of recent groundbreaking advances made possible by diffusion models~\cite{ho2020denoising, song2019generative}, there has been a surge of interest in extending these techniques to the 3D motion domain. The recent MotionDiffuse~\cite{zhang2022motiondiffuse} is regarded as the pioneering diffusion-based framework for text-driven motion generation. Similar to MotionDiffuse, the concurrent MDM~\cite{tevet2022human} and FLAME~\cite{kim2022flame} integrate diffusion models and pre-trained language models such as CLIP~\cite{radford2021learning} and RoBERTa~\cite{liu2019roberta} for generating motion from natural language descriptions. In a parallel development, speech audio is considered a model input~\cite{zhang2023diffmotion, alexanderson2022listen} for gesture synthesis. Compared with gesture synthesis, dance generation is perhaps a more complex and challenging, but relatively under-explored field. \citet{alexanderson2022listen} pioneers diffusion models with Conformer~\cite{zhang2022music} for music to dance generation. 
EDGE~\cite{tseng2022edge} and Magic~\cite{li2022magic} are Transformer-based diffusion models. 
EDGE incorporates auxiliary losses to encourage physical realism. Our work explores diffusion-based motion transfer on both human locomotion and dance databases. 

One line of study focuses on manipulating partial body movements and styles. The modeling is hence not only about what contents are expressed by gestures but also how they are executed. HumanMAC~\cite{chen2023humanmac} achieves controllable prediction of any part of the body. FLAME~\cite{kim2022flame} and MDM~\cite{tevet2022human} enable body part editing both frame-wise and joint-wise by adapting diffusion inpainting to motion data. \citet{tevet2022human} design a multi-task architecture to unify human motion synthesis and stylization. \citet{alexanderson2022listen} control the style and strength of motion expression by guided diffusion~\cite{dhariwal2021diffusion}. \citet{yin2023controllable} integrate multimodal transformer and autoregressive diffusion models for controllable motion generation and reconstruction. These works often require joint training of models for body parts and styles while this is not desired in our work for the application of dual diffusion implicit bridges~\cite{su2022dual}.

\subsection{Motion Style Transfer}
\label{subsec: motion style Transfer}

The field of computer animation has long grappled with the challenge of motion style transfer, a process that entails transferring animation from a source style to a target style while retaining key content aspects, such as structure, timing, spatial relationships, etc.
Early research in motion style transfer has depended on handcrafted features~\cite{amaya1996emotion, unuma1995fourier, witkin1995motion}.
As style is an elusive concept to define precisely, contemporary studies tend to endorse data-driven methodologies for feature extraction. 
Typical models utilized for style transfer include CycleGAN~\cite{dong2020adult2child}, AdaIN~\cite{aberman2020unpaired}, autoregressive flows~\cite{wen2021autoregressive}, and etc. 
Additionally, some research concentrates on addressing real-time style transfer~\cite{xia2015realtime, smith2019efficient, mason2022real}. Nonetheless, it is essential to highlight that these studies focus on relatively simplistic human motions, such as exercise and locomotion, where stylistic variation is restricted. CycleDance~\cite{yin2023dance} and StarDance~\cite{yin2023multimodal} address the transfer of dance movements that exhibit a substantial degree of complexity in terms of postures, transitions, rhythms, and artistic styles. However, CycleDance and StarDance suffer from severe drawbacks in rapid adaptation to alternative domains. As diffusion models advance, \citet{alexanderson2022listen} demonstrate the ability to control the style and intensity of dance motion expression using a classifier-free guided diffusion model. \citet{raab2023single} propose the diffusion-based SinMDM to transfer the style of a given reference motion to learned motion motifs. Our proposed framework explores diffusion models with manifold constraints to facilitate the transfer of dance style. The process of style transfer hinges on the diffusion models independently trained within each respective domain. This method overcomes the adaptability limitations of CycleDance and StarDance. We further boost the performance of style transfer by imposing keyframe manifold constraints. 

\section{Method}
\label{sec: method}
In this section, we formulate the problem and provide preliminaries of diffusion models and Dual Diffusion Implicit Bridges (DDIBs). 
After this, the proposed system with Keyframe Manifold Constraint Gradients (KMCGs) is presented. 

\begin{figure}[ht!]
\centering
  \includegraphics[width=0.99\linewidth]{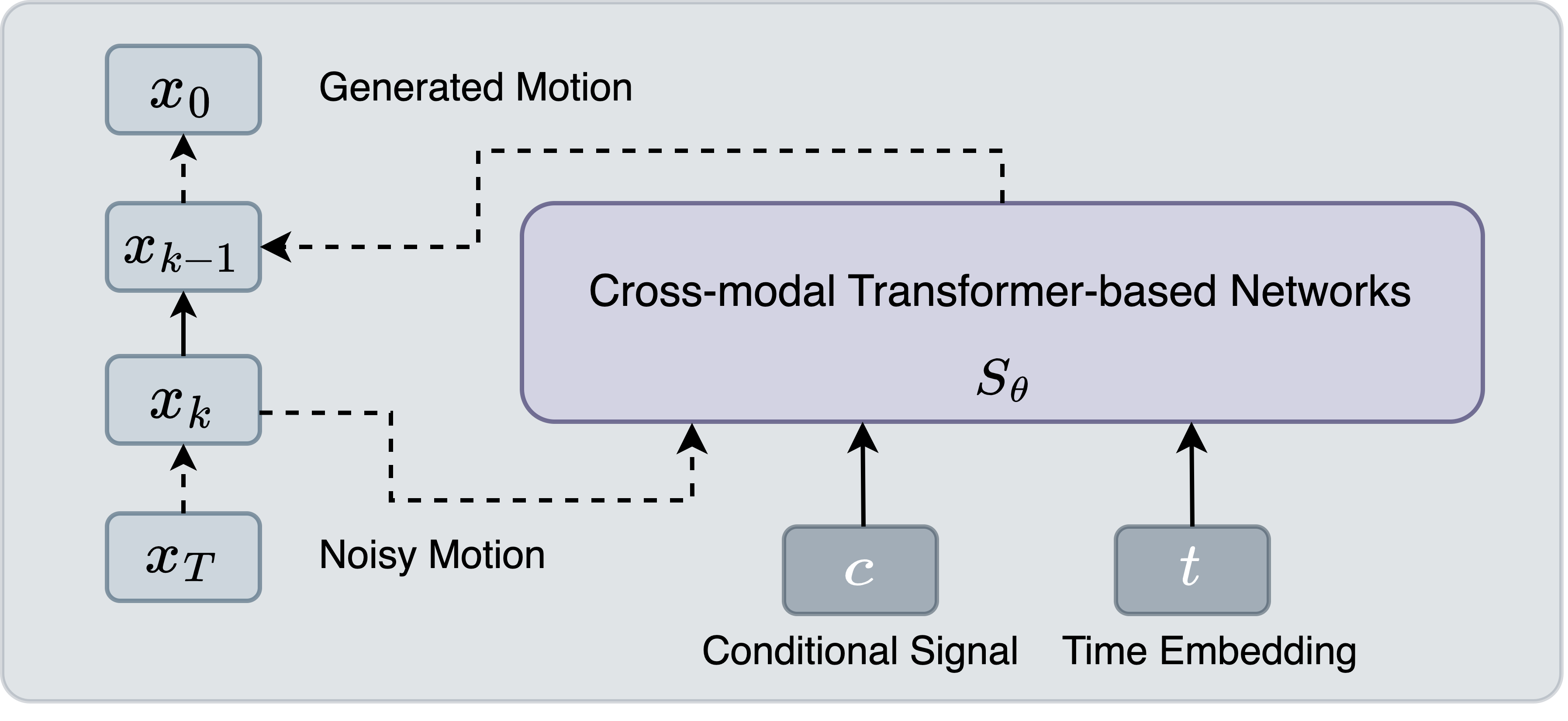}
  \caption{Architecture overview of diffusion models, conditional information acts as cross-attention context. The diffusion model takes noisy sequences and produces the estimated motion sequences. }
  \label{fig: model}
\end{figure}

\begin{figure}[t]
\centering
  \includegraphics[width=0.99\linewidth]{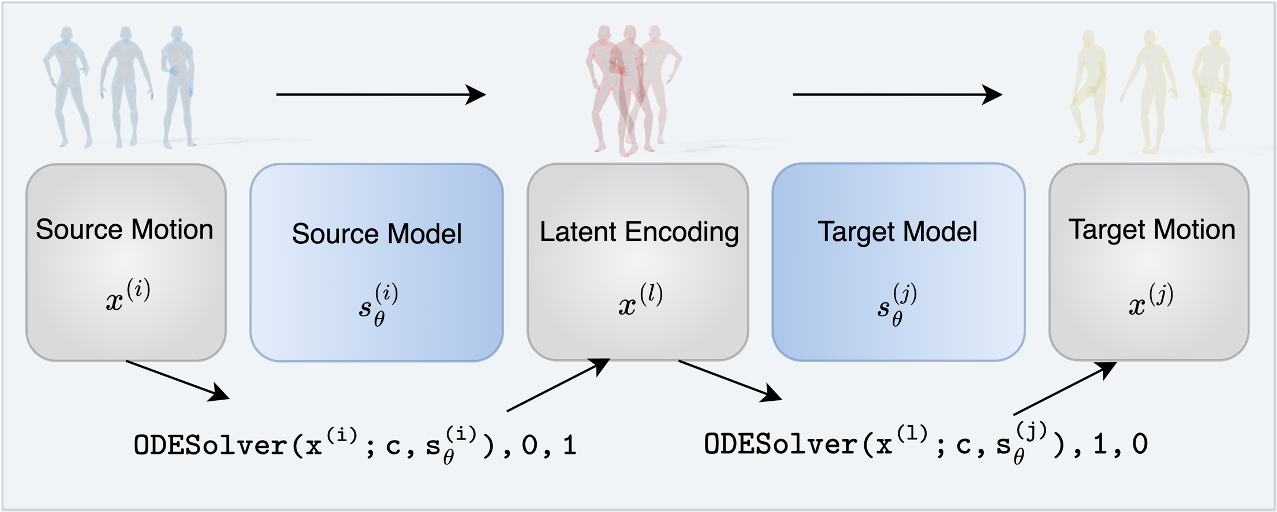}
  \caption{DDIBs leverage two models for style transfer. The source model converts the source sample from the source domain into the latent domain. The target model reverses the latent encoding to the target domain. }
  \label{fig:ddib}
\end{figure}

\subsection{Problem Formulation}
\label{subsec: problem formulation}

Our study aims to develop a translation method across multiple motion domains, denoted as source domain $D^{(i)}$ and target domain $D^{(j)}$, without relying on jointly training over multiple data domains. In our scenario, we focus on conditional motion sequences. Given a sample sequence $\bm{x}^{(i)}$ in the source domain and a conditional signal $\bm{c}$, our purpose is to transfer this sequence to the target domain, resulting in a sample sequence $\bm{x}^{(j)}$ with the style of the target domain while striving to retain the content of the source domain. 

\subsection{Dual Diffusion Motion Transfer}
To tackle the problem described above, we employ a strategy inspired by DDIB which use two separately trained Denoising Diffusion Implicit Models (DDIMs)~\cite{song2020denoising} for image-to-image translation. With DDIMs, DDIBs achieve exact cycle consistency. We further boost the transfer performance by imposing keyframe context from the source motion as a manifold constraint.

\begin{figure*}[ht]
\centering
  \includegraphics[width=0.88\textwidth]{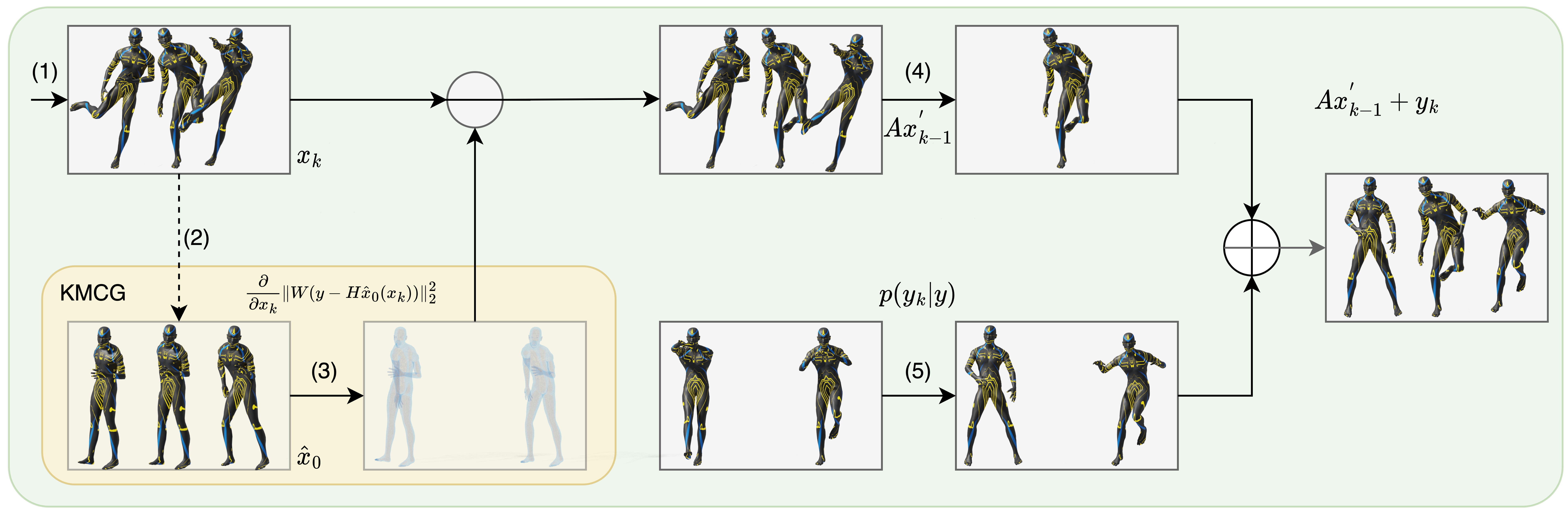}
  \caption{Visual schematic of the Keyframe Manifold Constraint Gradient (KMCGs) correction step. (1) unconditional reverse diffusion generates $\bm{x}_0$. (2) maps the noisy $\bm{x}_k$ to generate $\hat{\bm{x}}_0$. (3) the manifold constraint gradient $\frac{\partial}{\partial \bm{x}_k} \| \bm{W} (\bm{y} - \bm{H} \hat{\bm{x}}_0(\bm{x}_k)) \|_2^2$  is applied to constraint on the context manifold. (4) Takes the orthogonal complement. (5) Samples from $p(\bm{y}_k \| \bm{y})$, then combines $\bm{Ax}_{k-1}^{'}$ and $\bm{y}_k$ to further refine the motion transfer. }
  \label{fig: mcg}
\end{figure*}

\subsubsection{Deep Diffusion Models. }
Continuous-time diffusion processes $\{\bm{x}(t), t \in [0, 1] \}$ of diffusion models are defined as forward Stochastic Differential Equations (SDEs)~\cite{song2020score},  \begin{equation}
d\bm{x} = \bm{f}(\bm{x}, {t}) d{t} + g({t})d\bm{w},\label{eqn:forward_sde}
\end{equation}
where $\bm{w}$ is the standard Wiener process running backward in time, $f(\bm{x}, t)$ is the drift term, $g(t)$ is the scalar diffusion coefficient.
The backward-time SDEs of Eqn. \eqref{eqn:forward_sde} are 
\begin{equation}
d\bm{x} = [ \bm{f}(\bm{x}, {t})  - g^{2} \nabla_{\bm{x}} \log p_{\it{t}}(\bm{x})] \, d{t} + g({t}) \, d\bm{w},
\end{equation}
where $\nabla_{\bm{x}}\log p_t(\bm{x})$ is the score function of the noise perturbed data distribution at time $t$. The actual implementations of diffusion models are sample discrete times. The time horizon $t \in [0, 1]$ is split up into $T$ discretization segments as $\{\bm{x}_{{k}}\}_{k=0}^{T}$, where $k$ is an integer ranging from $0$ to $T$. The forward and backward diffusion process can be defined as:
\begin{equation}
\bm{x}_{{k}} = {a}_{{k}}\bm{x}_0 + {b}_{k}z,
\end{equation}
\begin{equation}
\label{eqa}
\bm{x}_{{k}-{1}} = {f}(\bm{x}_{{k}}, \bm{s}_\theta) + {g}(\bm{x}_k)z.
\end{equation}

We base our diffusion model on the EDGE architecture~\cite{tseng2022edge} to generate human motion contingent upon conditional signals. This architecture utilizes a transformer-based diffusion model that accepts conditional feature vectors as input. It then generates corresponding motion sequences as depicted in Figure~\ref{fig: model}. The model incorporates a cross-attention mechanism, following~\cite{saharia2022photorealistic}. We optimize the $\theta$-parameterized score networks $\bm{s}_\theta$ with paired conditional signal $\bm{c}$. The objective function is simplified as:
\begin{equation}
\mathbb{E}_{\bm{x}, {k}}\left\|\bm{x} - \bm{s}_{\theta}(\bm{x}_{{k}}, {k}, \bm{c}) \right\|^2_2.
\end{equation}

\subsubsection{Dual Diffusion Implicit Bridges. }

Unlike GAN-based approaches for style transfer~\cite{yin2023dance, yin2023multimodal}, which do not inherently ensure cycle consistency, the DDIBs~\cite{su2022dual} leverage the connections between score-based generative models~\cite{song2020score} and the Schrödinger Bridges Problem~\cite{chen2016relation}. To optimize cycle consistency across two domains, GAN-based methods require additional components during training, necessitating simultaneous access to multiple domains. This imposes constraints on both the system scalability and data privacy. In contrast, DDIBs establish deterministic bridges between distributions, operating as a form of entropy-regularized optimal transport. Consequently, cycle consistency is ensured up to the discretization errors of the ODE solvers, avoiding simultaneous access to multiple domains for consistency. 

The DDIBs-based strategy is depicted in Figure \ref{fig:ddib} and Algorithm \ref{alg:ddib}. The motion transfer process involves encoding source motion $\bm{x}^{(i)}$ and conditional signal $\bm{c}$ with the source diffusion model $\bm{s}_{\theta}^{(i)}$ to latent encoding $\bm{x}^{(l)}$ at the end time $t=1$, and then decoding the latent encoding using the target model $\bm{s}_{\theta}^{(j)}$ to construct target motion $\bm{x}^{(j)}$ at time $t=0$. Both steps are defined via ODEs. We then employ ODE solvers to solve the ODE and construct $\bm{x}^{(j)}$ at different times. In our experiment, we adopt DDIMs as the ODE solver. DDIMs generalize DDPMs via a class of non-Markovian diffusion processes that lead to the same training objective. 

\begin{algorithm}[t]
\caption{Motion style transfer with DDIBs}\label{alg:ddib}
\begin{algorithmic}
\State \textbf{Input:} source motion sample $\bm{x}^{(i)}{\sim}p_D^{(i)}$, conditional signal $\bm{c}$, source model $\bm{s}_{\theta}^{(i)}$, and target model $\bm{s}_{\theta}^{(j)}$
\State \textbf{Transfer:} 
\State source to latent: $\bm{x}^{(l)} = $ ODESolver$(\bm{x}^{(i)}; \bm{c}, \bm{s}_{\theta}^{(i)}, 0, 1)$
\State latent to target: $\bm{x}^{(j)} = $ ODESolver$(\bm{x}^{(l)}; \bm{c}, \bm{s}_{\theta}^{(j)}, 1, 0)$
\State \textbf{Output:} $\bm{x}^{(j)}$, transferred motion in target domain. 
\end{algorithmic}
\end{algorithm}

The trained diffusion models in DDIBs can be regarded as a summary of the domains of datasets. When translating, these models strive to generate motions in the target domain that is closest to the source motion in terms of optimal transport distances. This process is both a strength and a limitation of DDIBs. In image translation, when source and target domains share similarities, DDIBs typically succeed in identifying correct content. A similar phenomenon is observed in human locomotion domains, as patterns of locomotion across different domains are similar. DDIBs generally transfer correctly on human gaits. However, when datasets show less similarity  (e.g., birds and dogs), DDIBs may struggle to produce translation results that accurately retain the postures. In human motion, dance movements present a greater complexity and lesser similarity. Consequently, when employing DDIBs for dance style transfer, the system sometimes fails to preserve the content of the movements while transferring the style.

\subsubsection{Content Constrained Style Transfer. }
\label{sec: kmcg}

We propose incorporating additional corrective content to address the limitations of DDIBs discussed above:
\begin{equation}
\bm{y}=\bm{Hx}^i+\epsilon,
\end{equation}
where $\bm{y}$ is the measured content for motion correction, and 
$\epsilon$ is the noise in the measurement. In our implementation, we extract keyframes from the source motion as corrective content to retain the postures from the source motion in the target motion. The keyframes are selected based on the acceleration of body joints. 

The most straightforward method to incorporate additional information is explicitly adding these frames as part of model contexts. However, this approach has obvious drawbacks. It necessitates retraining the model, which negatively impacts the scalability of systems. Moreover, the inclusion of original motion frames as high-dimensional context can exacerbate the training stability. Nonetheless, we suggest this as an alternative way of enforcing domain information. A comparison to our proposed strategy can be found in the experimental evaluation section. 

To avoid extra training, we use an additional corrective term by manifold constraints~\cite{chung2022improving}. We take the keyframes from the source motion sample as a constraint, which leads to the defined keyframe manifold constrained gradients (KMCGs).
This term can synergistically operate with previous solvers to steer the reverse diffusion process closer to the context manifold. 
The reverse diffusion process in equation \ref{eqa} can be replaced by:
\begin{equation}
\bm{x}^{'}_{k-1} = \bm{f}(\bm{x}_{k}, s_\theta) - \alpha \frac{\partial}{\partial \bm{x}_k} \| (\bm{y} - \bm{H} \hat{\bm{x}}_0(\bm{x}_k)) \|_2^2 + {g}(\bm{x}_k)\bm{z},
\end{equation}
\begin{equation}
\bm{x}_{k-1} = \bm{A}\bm{x}^{'}_{{k}-1} + \bm{b}_{k},
\end{equation}
where $\bm{\hat{x}}_0$ is estimated $\bm{x}_0$. $\alpha$ depends on the noise covariance. Other parameters are defined by $\bm{A}=\bm{I}-\bm{H}^T\bm{H}$, $\bm{b}_k=\bm{H}^T\bm{y}$. 
Here we omit the target domain index $j$ in the reverse process for the notation brevity. 

Although keyframes in the target domain are not available in style translation problem formulation, we found that the keyframe context in the source motion as a manifold constraint can also significantly improve the performance of dance style transfer. 
We illustrate our scheme in Figure~\ref{fig: mcg}. The gradient term $\frac{\partial}{\partial \bm{x}_k} \| \bm{W} (\bm{y} - \bm{H} \hat{\bm{x}}_0(\bm{x}_k)) \|_2^2$ incorporates the information of $\bm{y}$ so the gradient of corrective term stays on the context manifold.

\section{Experiments}
To validate the capabilities of our motion style transfer system, we are eager to answer the questions: 1) can the system achieve scalable training? 2) can KMCGs enhance content preservation? 3) will the introduction of KMCGs compromise the strength of style transfer? In what follows, we present both objective and subjective evaluations of the system on domain standard human motion datasets with results favouring the proposed system and KMCGs.
\label{sec: experiments}

\subsection{Experimental Setting and Datasets}
\subsubsection{Experimental Setting. }
We compare our proposed system with KMCGs (DDIBs-gradient) to the baseline DDIBs-vanilla. For a fair comparison, we also add another ablation setting, DDIBs-explicit, which also has a direct access to source domain keyframes but incoporated through cross-attention. To remain indpendent training, DDIBs-explicit trains each style model with its own keyframes.  
To evaluate the system performance, we quantitatively assess the transfer strength and content preservation based on two probabilistic divergence-based metrics. The concrete metrics are presented below. Qualitative user studies are also important in the evaluation of generative models as well. As a complementary, we performed a subjective evaluation in the form of online survey to evaluate human perceived quality in terms of naturalness and content preservation. 

\subsubsection{Datasets. }
\label{subsubsec: datasets}

We evaluate our system on the 100STYLE~\cite{mason2022real}  locomotion database and the AIST++~\cite{tsuchida2019aist} dance database. We downsample both motion datasets to 30 fps and use 150-frame clips for experiments. 

For the 100STYLE dataset, we employ forward walking movements that encompass 100 diverse styles, such as angry, neutral, stiff, etc. We transform the motion context into a form that includes 24 body joints with 3D rotations (72-dimensional vector), supplemented with frame-wise delta translation and delta rotation of the root joint (3-dimensional feature) as control signals.

The AIST++ dataset includes ten dance genres: break, pop, locking, waacking, middle hip-hop, LA hip-hop, house, krump, street-jazz, and ballet-jazz dance. 
We used the same pose representation as in~\cite{tseng2022edge} that represent dance as sequences of poses in 24-joint SMPL format~\cite{loper2015smpl} using the 6-DOF rotation representation and 4-dimensional binary foot contact label (151-dimensional vector). The music features are extracted with a frozen Jukebox model~\cite{dhariwal2020jukebox}, resulting in a 4800-dimensional feature. 

\subsection{Evaluations}
\label{subsec: objective evaluation}

\subsubsection{Cycle Consistency. }
A desirable feature of the motion style transfer system is the cycle consistency property, which means transforming a motion sequence from the source domain to the target domain, and then back to the source, will recover the original data point in the source domain. 

\begin{figure}[ht!]
\centering
  \includegraphics[width=0.95\linewidth]{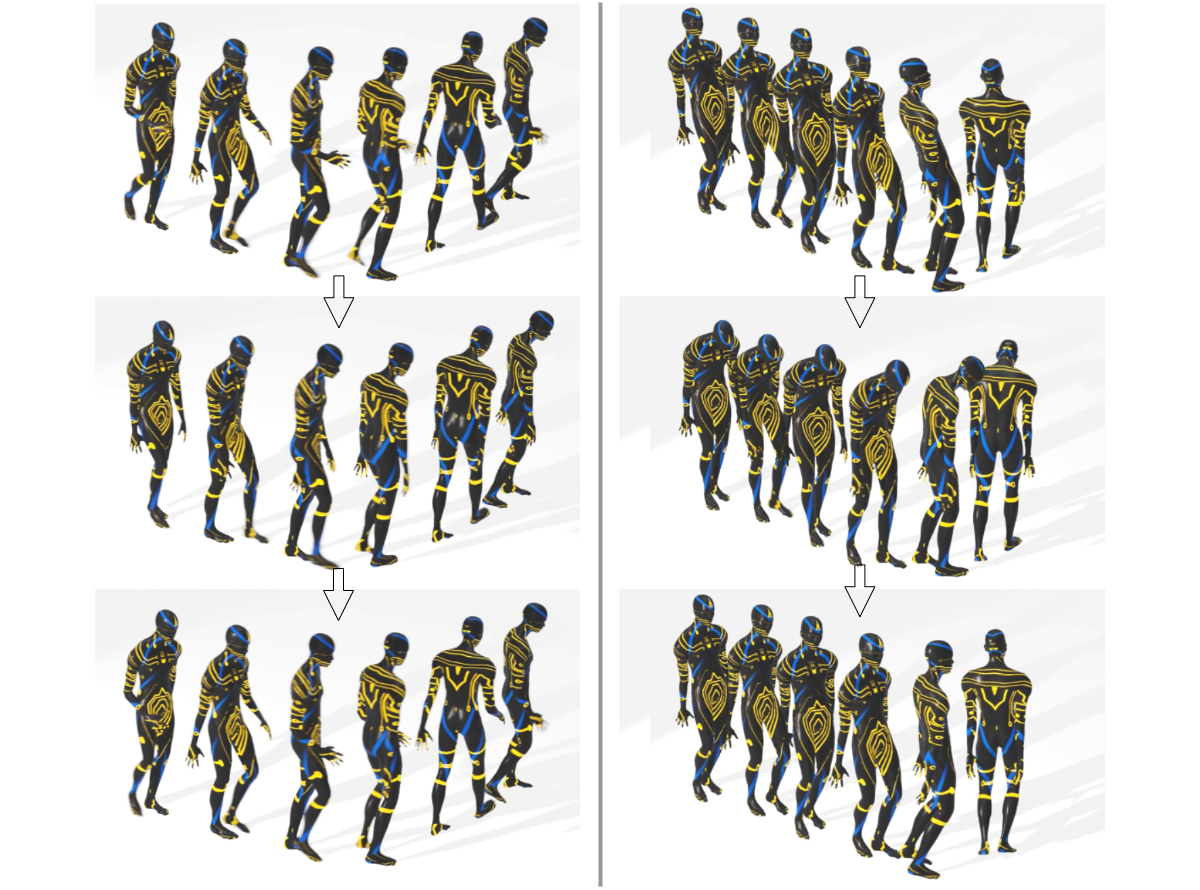}
  \caption{Cycle consistency: Translating the locomotion from the source style to the target style, and then back to the source style. Left: `proud' to `neutral' to `proud'. Right: `leanback' to `depressed' to `leanback'. }
  \label{fig: cyc}
\end{figure}

\begin{figure}[t]
\centering
  \includegraphics[width=0.8\linewidth]{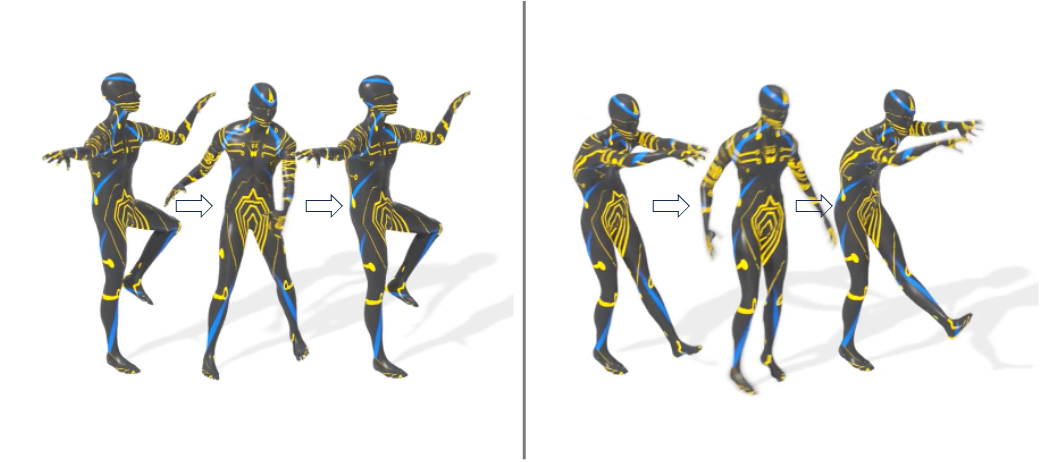}
  \caption{Cycle consistency: Translating the dance motion from the source style to the target style, and then back to the source style. Left: `breaking' to `hip-hop' to `breaking'. Right: `jazz-ballet' to `waacking' to `jazz-ballet'. }
  \label{fig: cyc2}
\end{figure}

Figure \ref{fig: cyc} illustrates the cycle consistency property guaranteed by DDIBs-based systems. It concerns the locomotion dataset.  Starting from the source domain, DDIBs first obtain the latent encodings and construct the motion in the target domain. Next, DDIBs do translations in the reverse direction, transforming the motion back to the latent and the source domain. After this round trip, motions are approximately mapped to their original patterns. This cycle consistency can also be observed in the dance dataset as shown in Figure \ref{fig: cyc2}. DDIBs-based transfer systems restore almost the exact same motion patterns as the original ones, which implies that noised latent encoding indeed carried information that may recover content. 
Table \ref{tab: cyc} reports quantitative results on cycle-consistent translation among both locomotion and dance styles, by DDIB-vanilla and DDIB-gradient. The reported values are negligibly small and endorse the cycle-consistent property in the human motion domain. 

\begin{table}[]
\centering
\caption{Cycle consistency of motion style transfer on locomotion and dance dataset. The numbers are the averaged $L_2$ distances between the original motion sequences and the motion after cycle translation. The motion data are standardized to have unit variance. }
\label{tab: cyc}
\resizebox{0.8\linewidth}{!}{
\begin{tabular}{lcc}
\hline
Method/dataset & 100STYLE &  AIST++\\
\hline
DDIBs-vanilla  & $0.0198 \pm 0.0083$ &  $0.0264 \pm 0.0035$\\
DDIBs-gradient &  $0.0192 \pm 0.0079$ & $0.0232 \pm 0.0032$ \\
\hline
\end{tabular}}
\end{table}

\subsubsection{Transfer Performance. }
The primary objective of this system is to transfer the motion style from a specified source domain to a particular target domain. We conduct assessments from both objective and subjective perspectives to ensure a comprehensive evaluation of the complex motion patterns common in dance. For the objective evaluation, we analyze the transfer strength and content preservation by 90 dance sequences for each style. Style transfer was performed on each ablated system and among all possible style pairs. Two metrics based on Fréchet distance are adopted~\cite{yin2023multimodal}, which is computed by Equation \ref{eq: fid}:
\begin{equation}
\label{eq: fid}
\begin{aligned}
    \text{FID} = \|\mu_{r} - \mu_{g}\|_{2}^{2} + \text{Tr}(\Sigma_{r} + \Sigma_{g} - 2\sqrt{\Sigma_{r} \Sigma_{g}}),
\end{aligned}
\end{equation}
where $(\mu_{r},\Sigma_{r})$ and $(\mu_{g},\Sigma_{g})$ are, respectively, the mean and the covariance matrix of the real and generated dance movement distribution. 

\begin{table}[htbp]
\centering
\caption{Quantitative objective evaluation: Fréchet Motion distance (FMD) and Fréchet Pose distance (FPD) are employed to evaluate the performance of the proposed transfer system and the baseline system. }
\resizebox{0.5\linewidth}{!}{
\label{tab: fid}
\begin{tabular}{lcc}
\hline
Method & FPD$\downarrow$ & FMD$\downarrow$ \\
\hline
StarDance & 0.4138 & 0.1816 \\
DDIBs-vanilla & 0.2904 & 0.1295 \\
DDIBs-explicit & 0.1748 & 0.1313 \\
DDIBs-gradient & 0.1208 & 0.1214 \\
\hline
\end{tabular}}
\end{table}

\begin{figure}[ht!]
  \includegraphics[width=\linewidth]{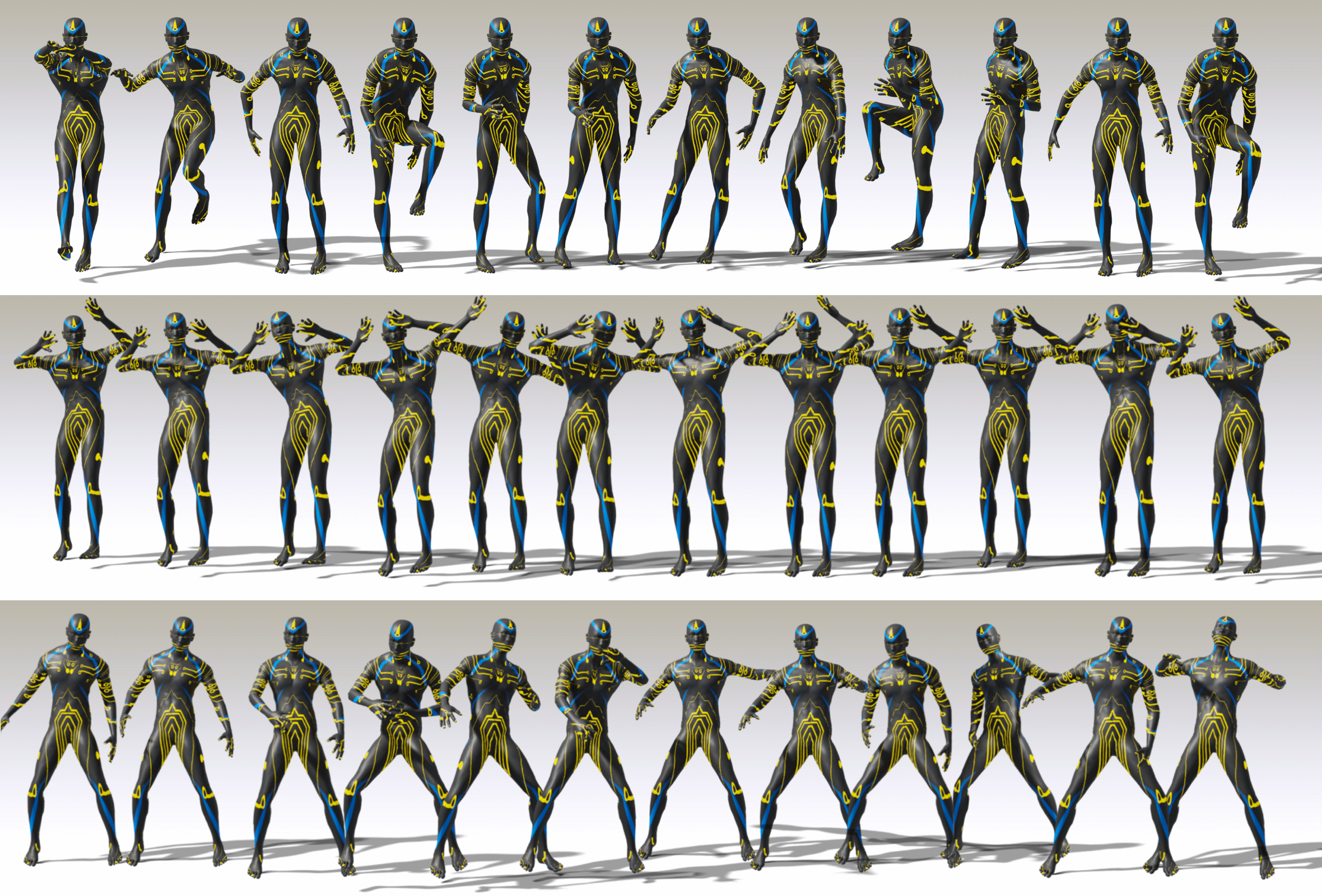}
  \caption{
  An example of transferring locking dance sequences (top) to krump dance using DDIBs-vanilla (middle) and DDIBs-gradient (bottom).  }
  \label{fig: compare}
\end{figure}

The Fréchet motion distance (FMD) quantifies transfer strength, which measures the extent that the motion is transferred from the source domain to the target domain. FMD calculates the distance between the true and generated dance motion distributions. The body joint acceleration and velocity are adopted as style-correlated features. 
The Fréchet pose distance (FPD) evaluates content preservation, which measures how well the salient poses of the source motion sequence are preserved after the transfer. The salient poses are detected by local maxima in joint acceleration and are normalized with respect to the hip-centric origin. 

The quantitative results for the proposed DDIBs-gradient and baselines across various dance style pairings are presented in Table \ref{tab: fid}. 
For StarDance, establishing a shared domain that bridges multiple styles presents a significant challenge. Other methods exhibit comparable performance on transfer strength, shows the introduced constraints do not compromise style transfer. However, in terms of content preservation, the baseline system DDIBs-vanilla encounters difficulties, as indicated by the higher FPD value relative to DDIBs-explicit and DDIBs-gradient.
The improvement in performance can be attributed to the explicit imposition of context information or through manifold constraint gradients.
An example of synthesized motion sequences showcasing the transfer of dance style from locking to krump is provided in Figure \ref{fig: compare}. The middle sequence is generated by DDIBs-vanilla, whereas the bottom sequence is created by DDIBs-gradient, which integrates keyframe context through manifold constraint gradients. By comparing the poses in each column, the DDIBs-gradient achieves a higher similarity in body postures to the source gesture in terms of body orientations and limb/body shapes, thus preserving more content.
Interestingly, DDIBs-gradient outperforms DDIBs-explicit. One possible explaination is that DDIBs-gradient gets to impose source domain related information across multiple intermediate diffusion steps. DDIBs-gradient is even more favoured for the possibility of directly using pre-trained models without requiring additional training and hyperparameter searching. 

\begin{table}[ht!]
\centering
\caption{Comparison of scalability between CycleDance and DDIBs-based methods.} 
\resizebox{0.99\linewidth}{!}{
\begin{tabular}{lcccc}
\hline
Dataset & Adult2Child & AIST+ & 100STYLE & PowerWash  \\ 
\hline
Number of Styles & 2 & 10 & 100 & 11080      \\
\hline
\multicolumn{5}{c}{Number of Models} \\ 
\hline
CycleDance & 2 & 90 & 9900 & 122901720  \\
DDIBs-based & 2 & 10 & 100 & 11080      \\
\hline
\end{tabular}
}
\label{tab2}
\end{table}

\subsubsection{Scalability and Data Privacy}
We further compare our proposed method with CycleDance concerning the scalability. 
Table~\ref{tab2} demonstrates that as the diversity of styles increases, our method scales the number of models linearly. In contrast, CycleDance surges the number of models quadratically, which would strain resources prohibitively. 
Consequently, CycleDance is appropriate for scenarios with a limited range of styles. As the scope of styles expands, our approach, characterized by its improved scalability, becomes increasingly advantageous.
In instances where data sensitivity is a priority due to privacy issues, such as with the Emopain@Home~\cite{olugbade2023emopain} dataset in rehabilitation therapy or the PowerWash~\cite{vuorre2023intensive} dataset in player behavior analysis. Our method secures data privacy by decoupling the training procedure. 

\subsubsection{User Study. }
\label{subsec: subjective evaluation}

To obtain a more comprehensive evaluation of our system and the baseline system, we conducted a user study in addition to the objective evaluation, asking participants to rate three aspects: motion naturalness, transfer strength, and content preservation. An online survey was performed to evaluate the transfer tasks. In this study, 21 participants between were recruited. Participants were between 24 and 35 years of age, 71.4\% male and 28.6\% female. 
We blend videos for each source and target dance sequence. During the survey, participants were presented with a source dance video clip followed by a generated target dance clip. To avoid potential order effects, the order of the target dance clips was randomly shuffled. Each target dance clip was generated either from the DDIBs-vanilla or DDIBs-gradient system. The participants were allowed to view the clips multiple times.

\begin{figure}[t!]
\centering
  \includegraphics[width=0.78\linewidth]{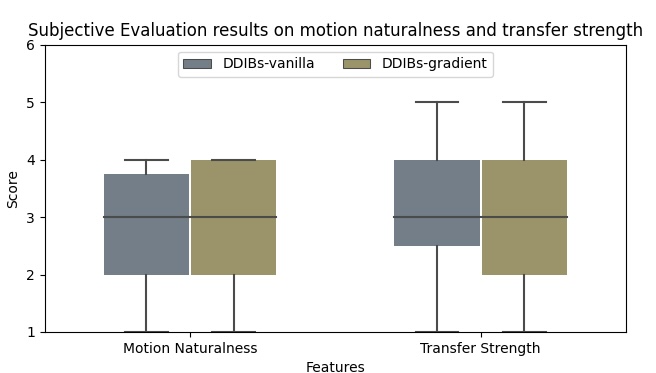}   
  \caption{
  Subjective evaluation results in motion naturalness and transfer strength. Equivalence tests were performed. 
  }
  \label{fig: sub1}
\end{figure}

For motion naturalness and transfer strength, the participants were asked to rate how much they agree with the statement that the motion is natural or transferred to the target style, using a 1-5 Likert scale, where ``1" means strongly disagree, and ``5" means strongly agree.  Figure \ref{fig: sub1} indicates that the performance on these two aspects is similar. 
We use the two one-sided tests to determine if the means of the two systems evaluations are equivalent. 
The equivalence margin was set at $\delta=0.5$. 
The differences are 0.2 and 0.35, which fall within the specified equivalence margin. 
The results showed there is no statistically significant difference between the means for the motion naturalness and transfer strength of DDIBs-vanilla and the DDIBs-gradient, which indicates that the constraints from source motion content will not lead to a decline in the motion naturalness and style transfer.  
\begin{figure}[t!]
\centering
  \includegraphics[width=0.78\linewidth]{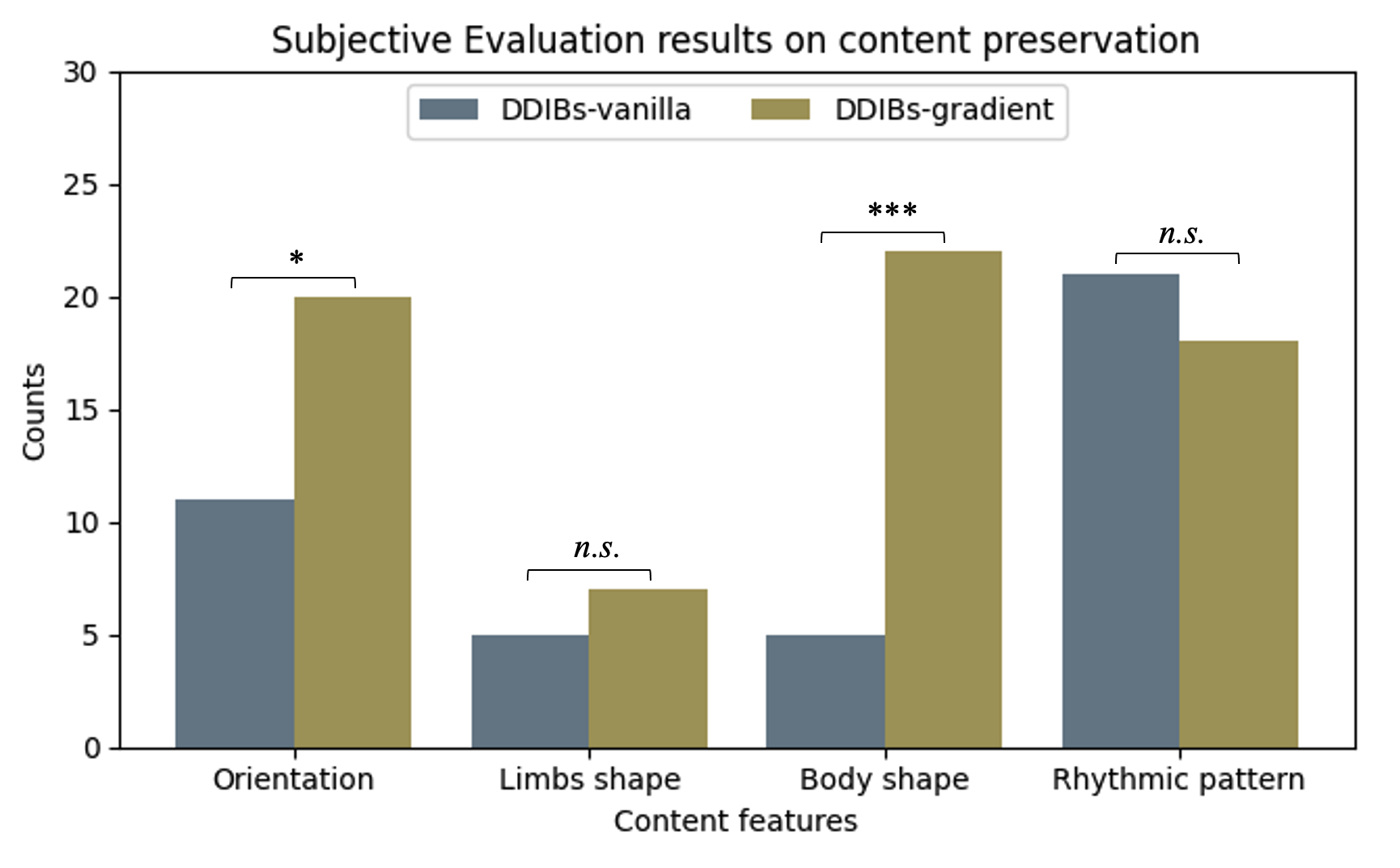}
  \caption{
  Subjective evaluation results in content preservation. 
  Statistical significance was determined using the 
  McNemar test ($*$ means $p<0.1$, $***$ means $p<0.001$, and $n.s.$ means $p>0.1$).
  }
  \label{fig: sub}
\end{figure}
For content preservation, the participants were asked to identify the features they believe are preserved between the source and the target motion, including orientations, limbs shape, body shape, and rhythmic patterns~\cite{newlove2004laban}. Figure \ref{fig: sub} presents the overall statistics for the four aspects. 
To assess the statistical significance of the differences between the DDIBs-vanilla and DDIBs-gradient systems, we conducted a McNemar test. The test showed no significant differences for `limb shape' ($p=0.25$), and `rhythmic' ($p=0.47$). However, there was a statistically significant difference for `orientations' ($p=0.07$) and `body shape' ($p=0.0000961$). Both systems scored high on `rhythmic', attributable to the DDIBs-based transfer system, which naturally preserved the rhythmic patterns due to its grounding in pre-trained music-conditioned diffusion models. In terms of preserving body trunk shape, the DDIBs-gradient outperformed DDIBs-vanilla because the posture information was provided as a constraint. As limb movements are more complex, participants reported that this content was not well preserved in either system.

\section{Conclusions}
\label{subsubsec: conclusions}
This study tackles the challenging task of motion style transfer. We propose a dual diffusion-based method with keyframe manifold constrain gradients. Our solution first converts the source motion into a latent encoding by pretrained source model, and then produces motion in the target domain by the pretrained target model. 
This framework addresses the scalability and data privacy issues associated with GAN-based motion style transfer systems. We improve the transfer performance on content preservation by guiding the transfer process with keyframe manifold constraint gradients. Extensive evaluations demonstrate the efficacy and superior performance of the proposed method.

\section{Acknowledgments}

\noindent We extend our gratitude to Ruibo Tu and Xuejiao Zhao for all the inspiring discussions and their valuable feedback. 

\bigskip

\noindent This research received partial support from the National Institute of Informatics (NII) in Tokyo. This work has been supported by the European Research Council (BIRD-884807) and H2020 EnTimeMent (no. 824160). 
This work benefited from access to the HPC resources provided by the Swedish National Infrastructure for Computing (SNIC), partially funded by the Swedish Research Council through grant agreement no. 2018-05973. 

\bibliography{aaai24}

\end{document}